\documentclass[10pt,twocolumn,letterpaper]{article}

\usepackage{cvpr}              

\usepackage{times}
\usepackage{epsfig}
\usepackage{graphicx}
\usepackage{amsmath}
\usepackage{amssymb}
\usepackage{xcolor}
\usepackage{caption}
\usepackage{subcaption}
\usepackage{lipsum}
\usepackage{booktabs}
\usepackage[accsupp]{axessibility}
\definecolor{mypurple}{HTML}{d000ff}

\usepackage[pagebackref,breaklinks,colorlinks]{hyperref}

\usepackage[capitalize]{cleveref}
\crefname{section}{Sec.}{Secs.}
\Crefname{section}{Section}{Sections}
\Crefname{table}{Table}{Tables}
\crefname{table}{Tab.}{Tabs.}

\def\cvprPaperID{30} 
\def\confName{CVPR}
\def\confYear{2022}

\begin{document}

\title{Towards Exemplar-Free Continual Learning in Vision Transformers:\\ an Account of Attention, Functional and Weight Regularization}
\author{Francesco Pelosin \thanks{Equal Contribution}\\
Ca' Foscari University\\
Venice, Italy\\
{\tt\small francesco.pelosin@unive.it}
\and
Saurav Jha\footnotemark[1]\\
University of New South Wales\\
Sydney, Australia\\
{\tt\small saurav.jha@unsw.edu.au}
\and
Andrea Torsello\\
Ca' Foscari University\\
Venice, Italy\\
{\tt\small torsello@unive.it}
\and
Bogdan Raducanu\\
Computer Vision Center\\
Barcelona, Spain\\
{\tt\small bogdan@cvc.uab.es}
\and
Joost van de Weijer\\ 
Computer Vision Center\\
Barcelona, Spain\\
{\tt\small joost@cvc.uab.es}
}

\maketitle

\begin{abstract}
In this paper, we investigate the continual learning of Vision Transformers (ViT) for the challenging exemplar-free scenario, with special focus on how to efficiently distill the knowledge of its crucial self-attention mechanism (SAM). Our work takes an initial step towards a surgical investigation of SAM for designing coherent continual learning methods in ViTs. We first carry out an evaluation of established continual learning regularization techniques. We then examine the effect of regularization when applied to two key enablers of SAM: (a) the contextualized embedding layers, for their ability to capture well-scaled representations with respect to the values, and (b) the prescaled attention maps, for carrying value-independent global contextual information. We depict the perks of each distilling strategy on two image recognition benchmarks (CIFAR100 and ImageNet-32) -- while (a) leads to a better overall accuracy, (b) helps enhance the rigidity by maintaining competitive performances. Furthermore, we identify the limitation imposed by the symmetric nature of regularization losses. To alleviate this, we propose an asymmetric variant and apply it to the pooled output distillation (POD) loss adapted for ViTs. Our experiments confirm that introducing asymmetry to POD boosts its plasticity while retaining stability across (a) and (b). Moreover, we acknowledge low forgetting measures for all the compared methods, indicating that ViTs might be naturally inclined continual learners.\footnote{Code will be made available at \url{https://github.com/srvCodes/continual_learning_with_vit}.}

\end{abstract}

\section{Introduction}
Transformers have shown excellent results for a wide range of language tasks \cite{DBLP:conf/nips/BrownMRSKDNSSAA20, DBLP:journals/tacl/RoySVG21} over the course of the last couple of years. Influenced by their initial results, Dosovitskiy \textit{et al.} \cite{DBLP:conf/iclr/DosovitskiyB0WZ21} proposed Vision Transformers (ViTs) as the first firm yet competitive application of transformers within the computer vision community.\footnote{By firmness, we refer to the non-reliance on convolutional operations.} ViTs' applications have since spanned a range of vision tasks, including, but not limited to image classification \cite{Touvron2021TrainingDI}, object recognition \cite{Liu2021SwinTH}, and image segmentation \cite{Wang2021EndtoEndVI}. The singlemost essential element of their architecture remains the self-attention mechanism (SAM) that allows the learning of long-range interdependence between the elements of a sequence (or patches of an image). Another  feature vital to their performance is the way they are pretrained in an often unsupervised or self-supervised manner over a large amount of data. This is then followed by the finetuning stage where they are adapted to a downstream task \cite{DBLP:conf/naacl/DevlinCLT19}.

For ViTs to be able to operate in real-world scenarios, they must exploit streaming data, \textit{i.e.,} sequential availability of training data for each task.\footnote{A task may encompass training data of one or more classes.} Storage limitations or privacy constraints further imply the restrictions on the storage of data from previous tasks. Task-incremental continual learning (CL) seeks to find solutions to such constraints by alleviating the event of \textit{catastrophic forgetting} - a phenomena where the network has a dramatic drop in performance on data from previous tasks. Several solutions have been proposed to address forgetting, including regularization \cite{imm, mas, si, laplace}, data replay \cite{tiny, mir, gem} and parameter isolation \cite{mallya2018packnet, rusu2016progressive, aljundi2017expert, dirichlet}. Nowadays most works on CL study recurrent \cite{Sodhani2020TowardTR, DBLP:conf/nips/ChiaroTB020} and convolutional neural networks (CNNs) \cite{ewc}. However, little has been done to investigate different CL settings in the domain of ViTs. We, therefore, mark the first step for the domain by considering the further restrictive setting of \emph{exemplar-free} CL with a zero overhead of storing any data from previous tasks. We consider this restriction for its real-world aptness to scenarios involving privacy regulations and/or data security considerations.

\begin{figure}[t]
\begin{center}
\centerline{\includegraphics[width=0.75\columnwidth]{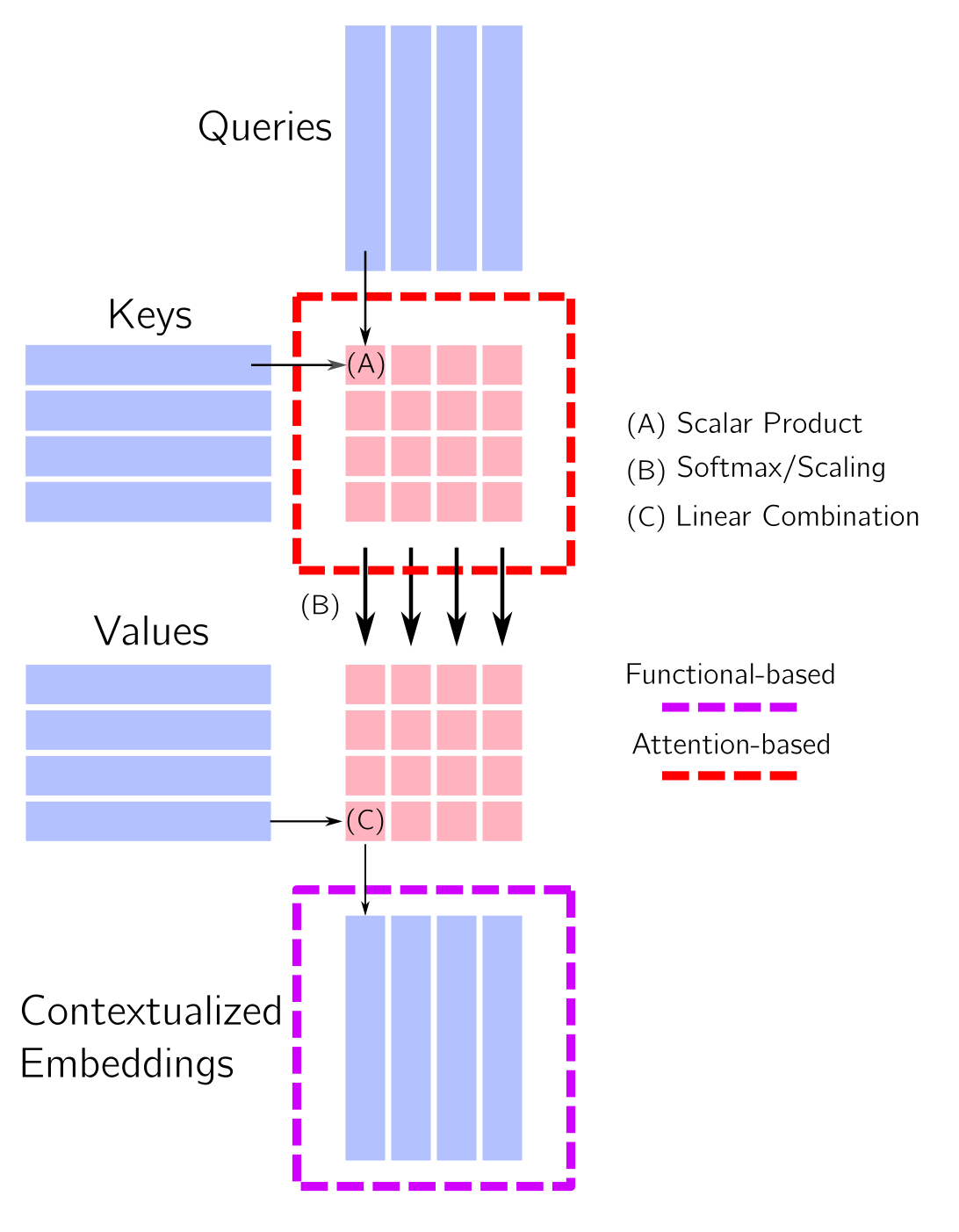}}
\caption{ Self-attention mechanism comprising a vision transformer encoder. We compare \textcolor{red}{Attention}-based approaches computed prior to the softmax operation and \textcolor{mypurple}{Functional}-based approaches computed on the contextualized embeddings. 
}
\label{fig:attention_functional}
\end{center}
\vskip -0.4in
\end{figure}

Given that regularization-based methods form one of the main techniques for exemplar-free CL, we consider an in-depth analysis of these for ViTs. Regularization-based techniques are mainly organized along two  branches: \emph{weight regularization} methods (such as EWC~\cite{ewc}, SI~\cite{si}, MAS~\cite{mas}) and \emph{functional regularization} methods ( such as LwF~\cite{lwf}, PODNET~\cite{podnet}). As discussed above, the architectural novelty of transformers lies in the SAM building a representation of a sequence by exhaustively learning relations among query-key pairs of its elements \cite{DBLP:conf/nips/VaswaniSPUJGKP17}. We show that for ViTs (and subsequently, all other architectures leveraging SAM), this property allows for a third form of regularization, which we coin as \emph{Attention Regularization} (see Figure \ref{fig:attention_functional}). We ground our idea in the hypothesis that when learning new tasks, the attention of the new model should still remain in the neighborhood of the attention of the previous model. As another contribution, we question the temporal symmetry currently applied to regularization losses; referring to the fact that they penalize the forgetting of previous knowledge and the acquiring of new knowledge equally (see Figure \ref{fig:asym_illustrate}). With the aim of countering forgetting while mitigating the loss of plasticity, we then propose an \textit{asymmetric} regularization loss that penalizes the loss of previous knowledge but not the acquiring of new knowledge. We index the major contributions of our work below:
\begin{itemize}
    \item We are the first to investigate continual learning in vision transformers in the more challenging \textit{exemplar-free} setting. We perform a full analysis of regularization techniques to counter catastrophic forgetting. 
    \item Given the distinct role of self-attention in modeling short and long-range dependencies \cite{Yang2021FocalSF}, we propose distilling the attention-level matrices of ViTs. Our findings show that such distillation offers accuracy scores on par with that of their more common functional counterpart while offering superior plasticity and forgetting. Motivated by the work of Douillard \textit{et al.} \cite{podnet}, we pool spatiality-induced attention distillation across our network layers.
    \item  We propose an asymmetric variant of functional and attention regularization which prevents forgetting while maintaining higher plasticity. Through our extensive experiments, we show that the proposed asymmetric loss surpasses its symmetric variant across a range of task incremental settings.
\end{itemize}

\section{Related Works}
Continual learning has been gaining contributions from the deep learning research community during the last few years~\cite{9349197,masana2020class}. 
In the following, we list the most prominent ones:

\begin{itemize}
\item Weight-based: these methods operate in the parameter space of the model through gradient updates. Elastic Weight Consolidation (EWC) \cite{ewc} and Synaptic Intelligence (SI) \cite{Zenke2017ContinualLT} are two widely used methods in this family with the former being
 probably, the most well-known. EWC uses Fisher information~\cite{pascanu2013revisiting} to identify the parameters important to individual tasks and penalizes their updates to preserve knowledge from older tasks. SI makes the neurons accumulate and exploit old task-specific knowledge to contrast forgetting.

\item Functional-based: these methods rely upon trading the plasticity for stability by training either the current (new) model on older data or vice-versa. Learning Without Forgetting (LWF) \cite{lwf} remains among the most widely known approaches in this family. It employs Knowledge Distillation \cite{kd} upon the logits of the network. 

\item Parameter Isolation-based: also known as architectural approaches, these methods tackle CF through a dynamic expansion of the network's parameters as the number of tasks grow. Among the well known methods in this family remain Progressive Neural Network (PNN) \cite{DBLP:journals/corr/RusuRDSKKPH16} followed by Dynamically Expandable Network (DEN) \cite{DBLP:conf/iclr/YoonYLH18} and Reinforced Continual Learning (RCL) \cite{DBLP:conf/nips/XuZ18}. 
\end{itemize}

The majority of the aforementioned works target CL in CNNs mainly due to their inductive bias allowing them to solve almost all problems that involve visual data. This can also be seen in several reviews \cite{Mai2022OnlineCL, biesialska-etal-2020-continual, 9349197, DBLP:journals/nn/ParisiKPKW19, DBLP:journals/nn/BelouadahPK21, DBLP:journals/ijon/MaiLJQKS22} reporting few approaches that consider architectures besides CNNs, despite the attempts to investigate CL in RNNs \cite{Sodhani2020TowardTR, DBLP:conf/nips/ChiaroTB020}.

Only recently have some works analyzed catastrophic forgetting in transformers. Among the earliest to do so remains that of Li \textit{et al.} \cite{li2022technical} proposing the continual learning with transformers (COLT) framework for object detection in autonomous driving scenarios. Using the Swin Transformer \cite{liu2021swin} as the backbone for a CascadeRCNN detector, the authors show that the extracted features generalize better to unseen domains hence achieving lesser
forgetting rates compared to ResNet50 and ResNet101 \cite{he2016deep} backbones. In case of ViTs,  Yu \textit{et al.} \cite{DBLP:journals/corr/abs-2112-06103} show that their vanilla counterparts are more prone to forgetting when trained from scratch. Alongside heavy augmentations, they employ a set of techniques to mitigate forgetting: (a) knowledge distillation, (b) balanced re-training of the head on exemplars (inspired by LUCIR \cite{Hou_2019_CVPR}), and (c) prepending a convolutional stem to improve  low-level feature extraction of ViTs.

 In their work studying the impact of model architectures in CL, Mirzadeh \textit{et al.} \cite{mirzadeh2022architecture} also experiment with ViTs in brief (with the rest of the work focusing mainly on CNNs). While they vary the number of attention heads of ViTs to show that this has little effect on the accuracy and forgetting scores, they further conclude that ViTs do offer more robustness to forgetting arising from distributional shifts when compared with their CNN-based counterparts with an equivalent number of parameters. The conclusion remains in line with previous works \cite{paul2021vision}.  Finally, Douillard \textit{et al.} \cite{DBLP:journals/corr/abs-2111-11326} attempt to overcome forgetting in ViTs through a parameter-isolation approach which dynamically expands the tokens processed by the last layer. For each task, they learn a new task-specific token per head. They then couple such approach with the usage of exemplars and knowledge distillation on backbone features. It is worth noting that these works rely either on pretrained feature extractors \cite{li2022technical} or rehearsal \cite{DBLP:journals/corr/abs-2112-06103, DBLP:journals/corr/abs-2111-11326} to defy forgetting. Thus the challenging scenario of \textit{exemplar-free} CL in ViTs remains unmarked.

\section{Methodology}

We start by shortly describing the two main existing regularization techniques for continual learning. We then propose attention regularization as an alternative approach tailored for ViTs. Lastly, we put forward an adaptation for functional and attention regularization which is designed to elevate plasticity while retaining stability levels.



\subsection{Functional and Weight Regularization}

\paragraph{Functional Regularization:} We include LwF \cite{lwf} in this component since it constitutes one of the most prominent, and perhaps the most widely used regularization method acting on data. The appealing property of LwF lies in the fact it is exemplar-free, \textit{i.e.}, it uses only the data of the current task and maintains only the model at task $t-1$ to exploit Knowledge Distillation \cite{kd}. Formally, LwF is defined as:
\begin{equation}
    \mathcal{L}_{\text{LwF}}(\theta)=\xi_{o} \mathcal{L}_{\mathrm{KD}}\left(Y_{o}, \hat{Y}_{o}\right)
\end{equation}
where $\mathcal{L}_{\mathrm{KD}}$ is the knowledge distillation loss incorporated to impose stability on the outputs, $\hat{Y}_{o}$ the predictions on the current task data from the old model and $\hat{Y}_{o}$ the ground truth of such data. $\xi_{o}$ remains the temperature annealing factor for softmax logits. To ensure better plasticity, $\mathcal{L}_{\mathrm{KD}}$ is often combined with the standard cross entropy loss $\mathcal{L}_{\mathrm{CE}}$ calculated upon the new task examples.

Some works have investigated the usage of functional regularization at intermediate layers of the network~\cite{liu2020generative,yu2020semantic,podnet}. This can be generalized to ViTs by applying a regularization on the contextualized embeddings (see purple block in Figure~\ref{fig:attention_functional}). In this case, the regularization loss could be defined in terms of functional distillation (FD): 
\begin{equation}
    \mathcal{L}_{FD}=  \left\| \sum_{w=1}^{W} \sum_{h=1}^{H} z^{t-1}_{l, w, h}-\sum_{w=1}^{W} \sum_{h=1}^{H} z^{t}_{l, w, h} \right\|
\end{equation}
where $\left\| . \right\|$ denotes the L2-norm, $z^{(t-1)}_l$ refers to the $l$-th layer output of the model trained at task $t-1$, and W and H are the width and height of the SAM outputs, respectively. 


\paragraph{Weight Regularization:} These methods encourage the network to adapt to the current task data mainly by using those parameters of the network that are not considered important for previous tasks. As representative method we select EWC \cite{ewc}. EWC exploits second-order information to estimate the importance of parameters for the current task. The importance is approximated by the diagonal of the Fisher Information Matrix $F$:
\begin{equation}
\mathcal{L}_{\text{EWC}}(\theta)=\mathcal{L}_{X}(\theta)+\sum_{j} \frac{\lambda}{2} F_{j}\left(\theta_{j}-\theta_{Y, j}^{*}\right)^{2}
\end{equation}
where $\mathcal{L}_{X}(\theta)$ is the loss for task X, $\lambda$ the regularization strength, and $\theta_{Y, j}^{*}$ the optimal value of $j^{t h}$ parameter after having learned task Y. 

\subsection{Attention Regularization}
\label{sec:att_reg}
\paragraph{Self-Attention Mechanism:} 

The self-attention mechanism (SAM) \cite{DBLP:conf/nips/VaswaniSPUJGKP17} forms the core of Transformer-based models and can be defined as:

\begin{figure*}[tb]
\begin{center}
\centerline{\includegraphics[width=0.7\textwidth]{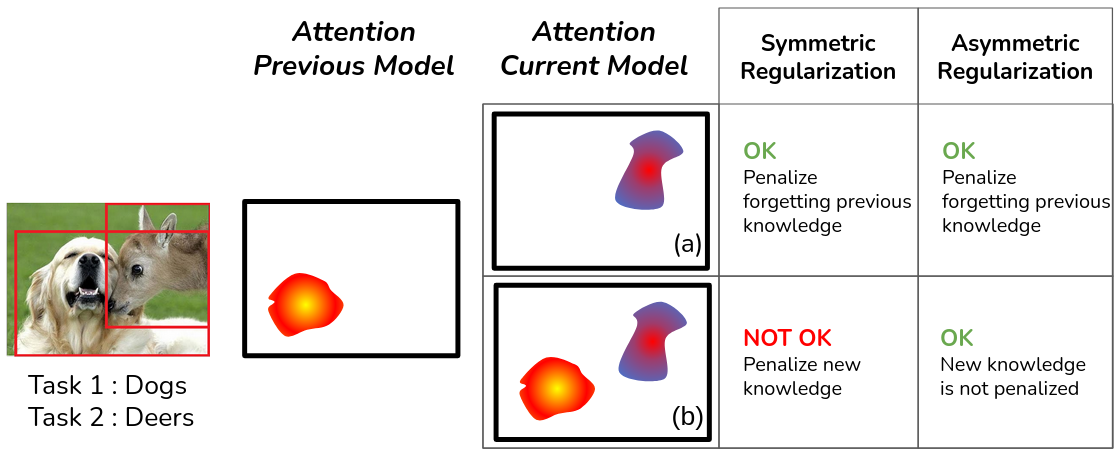}}
\caption{Visual illustration of the asymmetric loss with two generated attention maps (a) and (b) while training on task 2. In scenario (a), when previous knowledge is lost, both the symmetric and asymmetric regularization work correctly. However, in scenario (b), acquiring new knowledge is penalized by the symmetric loss but not by the asymmetric loss.}
\label{fig:asym_illustrate}
\end{center}
\vskip -0.4in
\end{figure*}
\begin{equation}
\mathbf{z} = \mathbf{softmax}\left ( \frac{\mathbf{QK}^T}{\sqrt{d_e}} \right )\mathbf{V}
\label{functional_var}
\end{equation}
where $\mathbf{Q,K}$, and $\mathbf{V}$ are respectively the projections of the Query, Key, and Values of the $\mathbb{R}^{d_e}$ input embeddings while $\mathbf{z}$ constitutes the new contextualized embeddings. Our novel attention-based regularization intervenes prior to the computation of the softmax operation of the standard self-attention mechanism as illustrated in Figure~\ref{fig:attention_functional}. 

In particular, given a ViT model at incremental step $t$ and  an SAM head $k$ of layer $l$, we define the prescaled attention matrix $\mathbf{A}^{t}_{k^{l}}$ prior to the softmax operation as:
\begin{equation}
    \mathbf{A}^{t}_{k^{l}} = \frac{\mathbf{QK}^T}{\sqrt{d_e}}
\end{equation} 
The attention matrix for time step $(t-1)$ can be similarly computed as $\mathbf{A}^{t-1}_{k^{l}}$. We employ this predecessor in the calculation of knowledge distillation in what follows.


\paragraph{Pooled Attention Distillation:}

Functional approaches leverage network's submodules typically to apply knowledge distillation \cite{kd}. 
When the regularization takes place in intermediate layers, the model can experience excessive stability, therefore loosing in plasticity abilities~\cite{podnet, liu2020generative, yu2020semantic}. Amongst these methods, PODNet \cite{podnet} clearly identifies the problem of excessive stability. We devise a regularization approach which instead of regularizing functional submodules targets attention maps, the core mechanisms of SAMs.

More formally, given the attention maps at steps $t$ and $(t-1)$, we define $\mathcal{L}_{\text {PAD}}\left(\mathbf{A}^{t-1}_{k^l}, \mathbf{A}^{t}_{k^l}\right)$ \cite{podnet} to be:

\begin{equation}
\label{eqn:spatial}
\mathcal{L}_{\text {PAD-width}}\left(\mathbf{A}^{t-1}_{k^l}, \mathbf{A}^{t}_{k^l}\right) + \mathcal{L}_{\text {PAD-height}}\left(\mathbf{A}^{t-1}_{k^l}, \mathbf{A}^{t}_{k^l}\right)
\end{equation}
\vskip -0.2in
\begin{align}
\label{eqn:pod_width_height_sym}
\begin{split}
\text{where, } \mathcal{L}_{\text{PAD-width}}\left(\mathbf{A}^{t-1}_{k^l}, \mathbf{A}^{t}_{k^l}\right) = \sum_{h=1}^{H} \mathcal{D}_W\left(\mathbf{A}^{t-1}_{k^l}, \mathbf{A}^{t}_{k^l}\right),
\\
\mathcal{L}_{\text{PAD-height}}\left(\mathbf{A}^{t-1}_{k^l}, \mathbf{A}^{t}_{k^l}\right) = \sum_{w=1}^{W} \mathcal{D}_H\left(\mathbf{A}^{t-1}_{k^l}, \mathbf{A}^{t}_{k^l}\right),\end{split}
\end{align}
\vskip -0.2in
\begin{equation}
\label{eqn:symdist}
   \mathcal{D}_X\left(\mathbf{A}^{t-1}_{k^l}, \mathbf{A}^{t}_{k^l}\right) =  \left\|\sum_{x=1}^{X} \mathbf{A}^{t-1}_{k^{l}_{w, h}}-\sum_{x=1}^{X} \mathbf{A}^{t}_{k^l_{w, h}} \right\|^2
\end{equation}
where, $W$ and $H$ indicate the width and height of the attention maps, and $\mathcal{D}_X(a,b)$ is the sum total of the distance measure between maps \textit{a} and \textit{b} along the arbitrary dimension \textit{X}. As shown in equation \ref{eqn:symdist}, the standard $\mathcal{L}_{\text{PAD}}$ uses the difference operator as the choice for $\mathcal{D}$. We point out the limitation of such  symmetric $\mathcal{D}$ and introduce, in the next section, the notion of \textit{asymmetry} into our distance measure.

As previously mentioned, Douilllard \textit{el al.} \cite{podnet} propose the pooled outputs distillation PODNet loss which leverages the symmetric Euclidean distance between the L2-normalized outputs of the convolutional layers of models at $t$ and $(t-1)$ after pooling them along specific dimension(s). They achieve the best results upon combining the pooling along the spatial width and  height axes which they term as the POD-spatial loss. Given the generic correspondence among the various pooling variants in their paper, our work is particularly influenced by POD-spatial as we pool attention maps of ViTs along two dimensions. In fact, throughout the experiments, we analyze this formulation when applied to the contextualized embeddings $\mathbf{z}$ resulting from a SAM operation. 
We would like to highlight that PAD differs from PODNet in two important factors: it is applied to the attention and not directly on the layer outputs, and its marginalization is not on the spatial dimensions due to the fact that $\mathbf{A}$ no longer encodes spatiality.


\subsection{Asymmetric Regularization}%
\label{sec:asym_reg}
The proposed attention regularization prevents forgetting of previous task by ensuring  
that the old attention maps are retained while the model learns to attend to new regions over tasks. However, the symmetric nature of $\mathcal{D}_X$  (with respect to the two attention maps) means that any differences between the older and the newly learned attention maps lead to increased loss values (see Equation \ref{eqn:pod_width_height_sym}). 
We agree that penalizing a loss in attention with respect to  previous knowledge is crucial in addressing forgetting. However, also penalizing a gain in attention for newly learned knowledge is undesirable and may actually hurt the performance over subsequently learned tasks. In other words, punishing additional attention can be counterproductive. As a result, we propose using an asymmetric variant of $\mathcal{D}_X$ that can better retain previous knowledge: 
\begin{equation}
\label{eqn:asym_dist}
\mathcal{D}_X\left(\mathbf{A}^{t-1}_{k^l}, \mathbf{A}^{t}_{k^l}\right) =  \left\|\mathcal{F}_{\text {asym}}\left(\sum_{x=1}^{X} \mathbf{A}^{t-1}_{k^l_{w, h}}- \sum_{x=1}^{X}\mathbf{A}^{t}_{k^l_{w, h}} \right) \right\|
\end{equation}
where, $\mathcal{F}_{\text{asym}}$ is as asymmetric function. We experimented with ReLU \cite{Nair2010RectifiedLU}, ELU \cite{DBLP:journals/corr/ClevertUH15} and Leaky ReLU \cite{Maas13rectifiernonlinearities} as choices for $\mathcal{F}_{\text{asym}}$ and found that in general, ReLU performed the best across our settings. The ReLU operation thus ensures that the new attention generated by the current model at task $t$ is not penalized while the attention present at task $t-1$ but missing in the current model $t$ is penalized. An illustration of the functioning of the new loss is provided in Figure~\ref{fig:asym_illustrate}.

Based on our choice for $\mathcal{D}_X$ from equations \ref{eqn:symdist} and \ref{eqn:asym_dist}, we classify our final PAD attention loss as asymmetric $\mathcal{L}_{\text{PAD-asym-att}}$ or symmetric $\mathcal{L}_{\text{PAD-sym-att}}$, respectively. Each of these losses are computed separately for each of the SAM head $k$ and model layer $lL$. The final asymmetric variant for attention regularization can be stated as:
\begin{equation}
\label{eqn:asym_final}
\begin{split}
   \mathcal{L}_{\text {PAD-asym-att}}(\mathbf{A}^{t-1}_{k^l}, \mathbf{A}^{t}_{k^l}) =\\  \frac{1}{L}\sum_{l=1}^{L} \frac{1}{K} \sum_{k=1}^{K}\mathcal{L}_{\text {PAD }}(\mathbf{A}^{t-1}_{k^l}, \mathbf{A}^{t}_{k^l})
   \end{split}
\end{equation}
where, $K$ and $L$ denote the total number of heads per layer and the total number of layers in the model, respectively. 
\begin{figure}[tb]
\begin{center}
\centerline{\includegraphics[width=0.6\columnwidth]{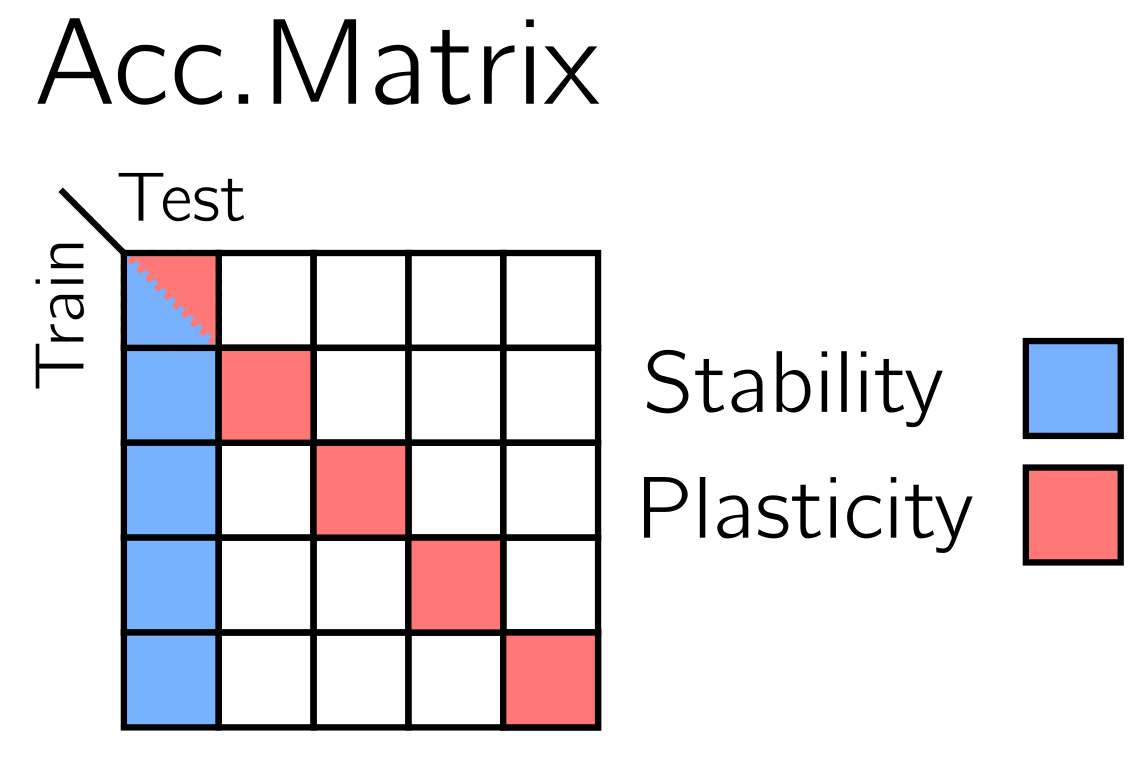}}
\caption{Illustration of a task accuracy matrix: we fix \textcolor{blue}{stability} to be the performance of the first task across the time steps while we define \textcolor{red}{plasticity} to be the performance at the current step.}
\label{fig:acc_mat}
\end{center}
\vskip -0.4in
\end{figure}
We now show that $\mathcal{L}_{\text{PAD-asym-att}}$ can accommodate functional distillation (FD) when applied to the contextualized global embeddings. To do so, we first replace the attention maps $A^t_{k^l}$ in Equation \ref{eqn:spatial}  by the contextualized vectors $z^t_{k^l}$ from Equation \ref{functional_var}. We denote the resulting functional counterpart of $\mathcal{L}_{PAD}$ by $\mathcal{L}_{FD}$. The final asymmetric variant for functional regularization can therefore be defined as: 
 
 \begin{equation}
\label{eqn:asym_final_functional}
\begin{split}
   \mathcal{L}_{\text {FD-asym-func}}(z^{t-1}_{k^l}, z^{t}_{k^l}) =\\  \frac{1}{L}\sum_{l=1}^{L} \frac{1}{K} \sum_{k=1}^{K}\mathcal{L}_{\text {FD }}(z^{t-1}_{k^l}, z^{t}_{k^l})
   \end{split}
\end{equation}

To avoid verbosity, we use the notation $\mathcal{L}_{\text {asym}}$ to symbolize the respective asymmetric losses from equations \ref{eqn:asym_final} and \ref{eqn:asym_final_functional}. Note that these equations can then be adapted to the symmetric variant $\mathcal{L}_{\text {sym}}$ without loss of generality.

\paragraph{Overall loss:} We augment the PAD losses from equations \ref{eqn:asym_final} and \ref{eqn:asym_final_functional} with knowledge distillation loss $\mathcal{L}_{LwF}$ \cite{lwf} and standard cross entropy loss $\mathcal{L}_{CE}$. The overall loss term thus takes the form:
\begin{figure*}[t]
\begin{center}
\centerline{\includegraphics[width=1\textwidth]{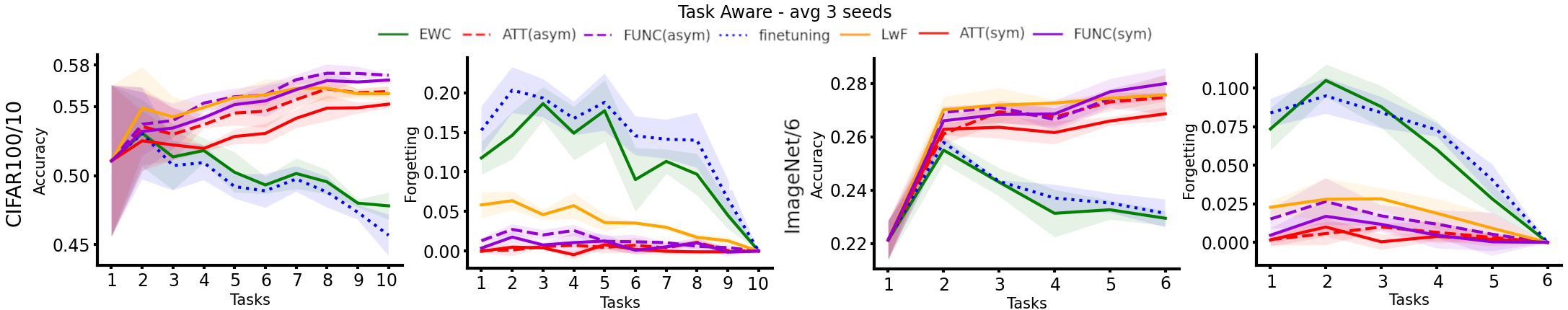}}
\caption{Mean and standard deviation of task-aware accuracy and forgetting scores for CIFAR100/10 and ImageNet/6 settings (over 3 random runs). Asymmetric approaches depict higher accuracy with respect to their symmetric counterparts. The low forgetting scores across all methods suggest an intrinsic forgetting resilience in vision transformer architectures. }
\label{fig:acc-forgetting}
\end{center}
\vskip -0.3in
\end{figure*}

\begin{equation}
    \label{eqn:total_loss}
    \mathcal{L} = \mu \mathcal{L}_{\text {(a)sym}} + \lambda \mathcal{L}_\mathrm{LwF}  + \mathcal{L}_{CE} 
\end{equation}
where $\mu$, $\lambda \in [0,1]$ are two hyperparameters regulating the respective contributions. Note that when $\mu = 0$, $\mathcal{L}$ degenerates to baseline finetuning for $\lambda = 0$ and to LwF for $\lambda = 1$.

\paragraph{Stability-Plasticity Curves:}
Several measures have been proposed in the CL literature to assess the performance of an incremental learner. Besides the standard incremental accuracy, Lopez-Paz \etal \cite{gem} introduce the notion of Backward Transfer (BWT) and Forward Transfer (FWT). BWT measures the ability of a system to propagate knowledge to past tasks, while FWT assesses the ability to generalize to future tasks. The CL community, however, still lacks consensus on a specific definition of the stability-plasticity dilemma. An elemental formulation for such quantification is thus desirable for allowing us to better grasp the trading capabilities of an incremental learner between acquiring new knowledge and discarding previous concepts. To this end, we introduce \emph{stability-plasticity curves} computed using task accuracy matrices. 

A task accuracy matrix $\mathbf{M}$ for an incremental learning setting composed of $T$ tasks is defined to be a $[0,1]^{T \times T}$ matrix, whose entries are the accuracies computed at each incremental step.\footnote{This calls for $\mathbf{M}$ to be lower trapezoidal.} For instance, $\mathbf{M}_{i,j}$ would constitute the test accuracy of task $j$ when the system is learning task $i$. Subsequently, the diagonal entries $\mathbf{M}_{i,i}$ give us the accuracies at the respective current tasks while the entries below the diagonal, \textit{i.e.,} $j<i$, give the performance of the model on past tasks. A visual depiction can be seen in Figure~\ref{fig:acc_mat}.

We define the stability to be the performance on the first experienced task at any given time and plasticity to be the ability of the model to adapt to the current task. Namely, these constitute the first column $\mathbf{M}_{:,0}$ and the diagonal of the matrix $diag(\mathbf{M})$. We employ the curves derived from these definitions to better dissect the stability-plasticity dilemma of the methods analyzed in our work.


\section{Experiments}
In this section, we compare regularization-based methods for \textit{exemplar-free} continual learning. We evaluate the newly proposed attention regularization and compare it with the existing functional (\emph{LwF}) and weight (\emph{EWC}) regularization methods. We then ablate the usefulness of the newly proposed asymmetric loss as well as the importance of pooling prior to applying the regularization.

\subsection{Experimental Setup}

\paragraph{Setting:}

For our experiments, we adopt the variation of ViTs introduced by Xiao \textit{et al.} \cite{early_conv}.  Here, the standard linear embedder of a ViT model is replaced by a smaller convolutional stem which helps build more resilient low-level features. Convolutional stems have previously been shown to improve performance and convergence speed in incremental learning settings \cite{DBLP:journals/corr/abs-2112-06103}. We therefore define our architecture to be a lightweight variation of a ViT-Base by setting $L=12$ layers, $K=12$ heads per layer and a $d_e = 192$ embedding size. The choice of a small embedding size has been made to speed up the training procedure and unlock the ability to handle larger batch sizes (1024 for our work).

We analyze our task-incremental setting on two widely used image recognition datasets - namely CIFAR100 and ImageNet-32 with 100, and 300 classes each. Both datasets host $32\times32$ images. On CIFAR100, we consider a split of 10 tasks (denoted as CIFAR100/10 setting) where each incremental task is composed of 10 disjoint set of classes. On ImageNet-32, we split 6 tasks with 50 disjoint set of classes each (denoted as ImageNet/6 setting).\footnote{Refer to Appendix \ref{tag_appendix} for experiments on additional settings.} 

Our total training epochs remain 200 (per task) for CIFAR100 and 100 for ImageNet32 with an initial learning rate of 0.01 and patience set of 20 epochs. We report our scores averaged over 3 random runs.
We apply a constant padding of size 4 across all our datasets. The train images are augmented using random crops of sizes $32\times32$ and random horizontal flips with a flipping probability of 50\%. For test images, we only apply center crops of sizes $32\times32$.

We compare the attention and functional symmetric and asymmetric versions of $\mathcal{L}_{\text {(a)sym}}$. Our basic functional and weight regularization approaches include  LwF \cite{lwf} and EWC \cite{ewc}, respectively. For all our experiments relying on PAD losses, we performed a hyperparemeter search (using equation \ref{eqn:total_loss}) for $\mu$ and $\lambda$ by varying each in the range $[0.5, 1.0]$ and found $\mu = \lambda = 1.0$ to perform reasonably well. We thus stick to these values unless otherwise specified. For the sake of readability, the functional and attentional variants of $\mathcal{L}_{\text{asym}}$ are indicated as FUNC(asym) and ATT(asym), respectively. Similarly, those corresponding  to $\mathcal{L}_{\text{sym}}$ are denoted by FUNC(sym) and ATT(sym). Note that all these four losses specify instantiations of equation \ref{eqn:total_loss}. The functional approaches regularize the contextualized embeddings while the attention approaches leverage prescaled attention maps  (see Figure~\ref{fig:attention_functional}).

\subsection{Results}
\label{sec:main_results}
We report accuracy as well as forgetting \cite{chaudhry2018riemannian} scores in task aware (taw) setting.\footnote{The corresponding task agnostic scores can be found in Figure \ref{fig:cifar100_tag_settings}, Appendix \ref{tag_appendix}.} We further report taw plasticity-stability curves (based on Figure \ref{fig:acc_mat}) to provide insights upon how well the different models trade the dilemma.

\begin{figure*}[t]
\begin{center}
\centerline{\includegraphics[width=\textwidth]{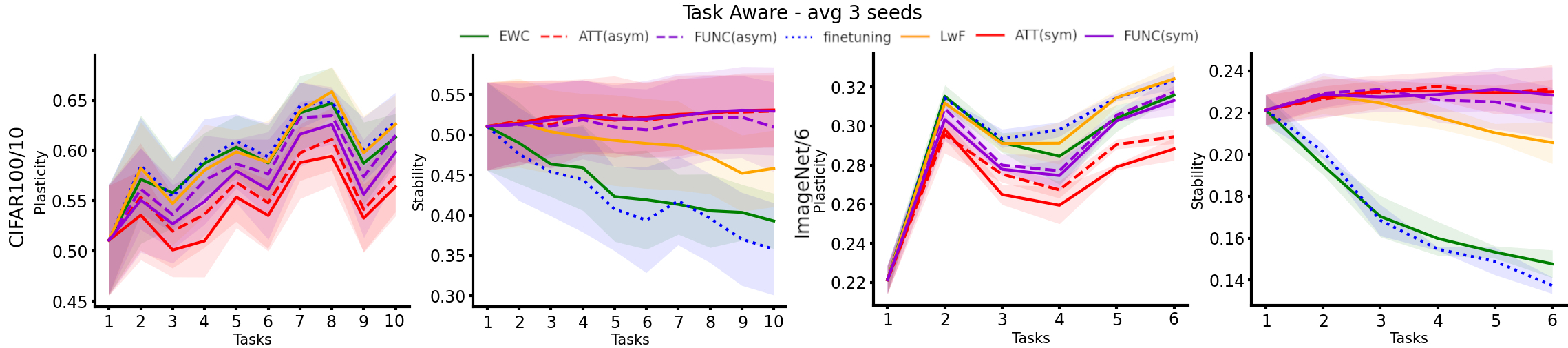}}
\caption{Mean and standard deviation of task-aware plasticity-stability scores for CIFAR100/10 and ImageNet/6 settings (over 3 random runs). Asymmetric approaches are more plastic compared to their symmetric counterparts while retaining competitive stability.}
\label{fig:cifar_stab_plast}
\end{center}
\vskip -0.3in
\end{figure*}

\paragraph{Accuracy and Forgetting:}
As seen in  Figure~\ref{fig:acc-forgetting}, all asymmetric approaches show better performances  with respect to their symmetric counterparts on CIFAR100/10 with ATT(asym) offering the best accuracy of 57.3\% on the last task. The trend continues for ImageNet/6 with an exception of asymmetric functional approach with an accuracy of 27.55\% falling behind its symmetric counterpart by 0.44\%. 
In general, the asymmetric and symmetric losses lead to improved accuracy scores with respect to other methods. Moreover, we observe that all the methods depict good forgetting resilience with their forgetting scores running at $\approx$0.01\% except for EWC. This suggests us that vision transformers are \textit{better incremental learners} but require more training  and tuning efforts to achieve reasonable accuracies. This remark remains in accordance with prior studies \cite{mirzadeh2022architecture, paul2021vision}. In the particular case of EWC, we observe poor compatibility in terms of accuracy as well as forgetting -- with the scores falling behind finetuning at times. We suspect that the method might be less suited for ViTs due to its reliance on exhaustive fisher information estimation.

\paragraph{Plasticity-stability trade off:}
We compare the dilemma for various methods in Figure~\ref{fig:cifar_stab_plast}. With no distillation, finetuning is prone to the worst trading of plasticity for stability. Meanwhile, our asymmetric losses can be seen to be more plastic with respect to their symmetric counterparts while depicting comparable stability scores. This confirms our hypothesis regarding the nature of the asymmetry keeping it from discarding older attention while favoring the integration of new attention at the same time. Although, LwF with a last task score of 47.74\% on CIFAR100/10 and 32.0\% on ImageNet/6, reports the best plasticity among our approaches, it clearly lags behind the pooling-based approaches at retaining stability. On the contrary, the (a)symmetric attention losses and the symmetric functional loss perform similar with a last task stability score of $\approx0.23\%$ on ImageNet/6 and $\approx53\%$ on CIFAR100/10. EWC shows good plasticity but virtually zero stability. This trend is in line with our previous remark on the limitation of EWC in Figure~\ref{fig:acc-forgetting}.



\begin{table}[]
\resizebox{\columnwidth}{!}{%
\begin{tabular}{@{}cccccc@{}}
\multicolumn{6}{c}{\textbf{CIFAR100/10  (taw)}}                                                                                                                                                                                                                                                                                                                                              \\ \midrule
\textbf{}                                                                    & \textit{\begin{tabular}[c]{@{}c@{}}FUNC(asym) \\ Spatial\end{tabular}} & \textit{\begin{tabular}[c]{@{}c@{}}FUNC(sym) \\ Spatial\end{tabular}} & \textit{\begin{tabular}[c]{@{}c@{}}FUNC(asym) \\ Intact\end{tabular}} & \textit{\begin{tabular}[c]{@{}c@{}}FUNC(sym) \\ Intact\end{tabular}} & \textit{LwF} \\
\textit{\begin{tabular}[c]{@{}c@{}}Average Incr.\\  Accuracy\end{tabular}} & \textbf{56.18\%}                                                               & 55.67\%                                                              & 54.43\%                                                             & 53.12\%                                                            & 54.81\%      \\
\textit{\begin{tabular}[c]{@{}c@{}}Last Task \\ Accuracy\end{tabular}}           & \textbf{57.26\%}                                                               & 56.92\%                                                              & 56.04\%                                                             & 54.59\%                                                            & 55.93\%      \\ \bottomrule
\end{tabular}
}
\caption{Comparison of intact (no pooling), spatial (pooling along width and height), and LwF.}
\label{tbl:loss_ablation}
\end{table}

\subsection{Ablation study}

Towards the end goal of evaluating the effectiveness of the best performing FD losses, we ablate the contribution of pooling on the CIFAR100/10 setting.
In particular, we consider distilling the contextualized embeddings when these are: (a) pooled along both dimensions, \textit{i.e.,}FUNC(asym/sym) Spatial (see Equation \ref{eqn:spatial}), and (b) not pooled at all, \textit{i.e.,} FUNC(asym/sym) Intact. Distilling the intact embedding vectors of the latter setting obviously implies enhanced stability over its pooled counterpart. Our standard accuracy and plasticity-stability measures across tasks can therefore be deemed redundant in this setting. As a consequence, we choose to compare the task-aware average incremental accuracy \cite{rebuffi2017icarl} and the last task accuracy across (a) and (b) while contrasting these with LwF as a strong baseline. As shown in Table \ref{tbl:loss_ablation}, we find that FUNC(asym) Spatial consistently performs the best on both the metrics (with a gain of  $>2\%$ over FUNC(sym) Intact and $>1\%$ over LwF in either metric). In general,  distilling the intact embeddings can be seen to be hurting the performance of the models as their accuracies drop below that of the baseline LwF. This suggests us that strong regularization effects such as that from the distillation of intact contextualized embeddings can often lead to ungenerous trading of stability for plasticity.



\section{Conclusion}

In this work, we adapted and analyzed several continual learning methods to counter forgetting in Vision Transformers mainly with the help of regularization. We then introduced a novel self-attention inspired regularization, based on the attention maps of self-attention mechanisms which we termed as Pooled Attention Distillation (PAD). Discerning its limitation at learning new attention, we devised its asymmetric version that avoids penalizing the addition of new knowledge in the model. We validated the superior plasticity of the asymmetric loss on several benchmarks. 

Besides the detailed comparison with a range of regularization approaches, \textit{i.e.}, functional (LwF), weight (EWC), and the proposed PAD regularization, we extended the application of PAD to the functional submodules of Vision Transformers. To this end, we investigated the functional distillation (FD) loss applied in the contextualized embeddings of ViTs.  The latter exploration led us to discover that the regularization of functional submodules can help achieve the best overall performances while the regularization of their attentional counterparts endow CL models with superior stability. Finally, we remarked the low forgetting scores of vision transformers across the incremental tasks and concluded that their enhanced generalization capabilities may offer them a natural inclination towards incremental learning.  By releasing our code, we hope to open the doors for future research along the direction of efficient continual learning with transformer-based architectures.

\section*{Acknowledgment}
We acknowledge the support from Huawei Kirin Solution, the Spanish Government funded project PID2019-104174GB-I00/ AEI / 10.13039/501100011033, and European Union's Horizon 2020 research and innovation programme under the Marie Sklodowska-Curie grant agreement number 777720 (CybSpeed). 




{\small
\bibliographystyle{ieee_fullname}
\bibliography{egbib.bib}

\begin{thebibliography}{10}\itemsep=-1pt

\bibitem{mas}
Rahaf Aljundi, Francesca Babiloni, Mohamed Elhoseiny, Marcus Rohrbach, and
  Tinne Tuytelaars.
\newblock Memory aware synapses: Learning what (not) to forget.
\newblock In {\em Proceedings of the European Conference on Computer Vision
  (ECCV)}, 2018.

\bibitem{mir}
Rahaf Aljundi, Eugene Belilovsky, Tinne Tuytelaars, Laurent Charlin, Massimo
  Caccia, Min Lin, and Lucas Page-Caccia.
\newblock Online continual learning with maximal interfered retrieval.
\newblock In {\em Advances in Neural Information Processing Systems 32}. 2019.

\bibitem{aljundi2017expert}
Rahaf Aljundi, Punarjay Chakravarty, and Tinne Tuytelaars.
\newblock Expert gate: Lifelong learning with a network of experts.
\newblock In {\em Proceedings of the IEEE Conference on Computer Vision and
  Pattern Recognition}, 2017.

\bibitem{DBLP:journals/nn/BelouadahPK21}
Eden Belouadah, Adrian Popescu, and Ioannis Kanellos.
\newblock A comprehensive study of class incremental learning algorithms for
  visual tasks.
\newblock {\em Neural Networks}, 2021.

\bibitem{biesialska-etal-2020-continual}
Magdalena Biesialska, Katarzyna Biesialska, and Marta~R. Costa-juss{\`a}.
\newblock Continual lifelong learning in natural language processing: A survey.
\newblock In {\em Proceedings of the 28th International Conference on
  Computational Linguistics}. International Committee on Computational
  Linguistics, 2020.

\bibitem{DBLP:conf/nips/BrownMRSKDNSSAA20}
Tom~B. Brown, Benjamin Mann, Nick Ryder, Melanie Subbiah, Jared Kaplan,
  Prafulla Dhariwal, Arvind Neelakantan, Pranav Shyam, Girish Sastry, Amanda
  Askell, Sandhini Agarwal, Ariel Herbert{-}Voss, Gretchen Krueger, Tom
  Henighan, Rewon Child, Aditya Ramesh, Daniel~M. Ziegler, Jeffrey Wu, Clemens
  Winter, Christopher Hesse, Mark Chen, Eric Sigler, Mateusz Litwin, Scott
  Gray, Benjamin Chess, Jack Clark, Christopher Berner, Sam McCandlish, Alec
  Radford, Ilya Sutskever, and Dario Amodei.
\newblock Language models are few-shot learners.
\newblock In Hugo Larochelle, Marc'Aurelio Ranzato, Raia Hadsell,
  Maria{-}Florina Balcan, and Hsuan{-}Tien Lin, editors, {\em Advances in
  Neural Information Processing Systems 33: Annual Conference on Neural
  Information Processing Systems 2020, NeurIPS 2020, December 6-12, 2020,
  virtual}, 2020.

\bibitem{chaudhry2018riemannian}
Arslan Chaudhry, Puneet~K Dokania, Thalaiyasingam Ajanthan, and Philip~HS Torr.
\newblock Riemannian walk for incremental learning: Understanding forgetting
  and intransigence.
\newblock In {\em Proceedings of the European Conference on Computer Vision
  (ECCV)}, 2018.

\bibitem{tiny}
Arslan Chaudhry, Marcus Rohrbach, Mohamed Elhoseiny, Thalaiyasingam Ajanthan,
  Puneet~K. Dokania, Philip H.~S. Torr, and Marc'Aurelio Ranzato.
\newblock On tiny episodic memories in continual learning, 2019.

\bibitem{DBLP:conf/nips/ChiaroTB020}
Riccardo~Del Chiaro, Bartlomiej Twardowski, Andrew~D. Bagdanov, and Joost
  van~de Weijer.
\newblock {RATT:} recurrent attention to transient tasks for continual image
  captioning.
\newblock In Hugo Larochelle, Marc'Aurelio Ranzato, Raia Hadsell,
  Maria{-}Florina Balcan, and Hsuan{-}Tien Lin, editors, {\em Advances in
  Neural Information Processing Systems 33: Annual Conference on Neural
  Information Processing Systems 2020, NeurIPS 2020, December 6-12, 2020,
  virtual}, 2020.

\bibitem{DBLP:journals/corr/ClevertUH15}
Djork{-}Arn{\'{e}} Clevert, Thomas Unterthiner, and Sepp Hochreiter.
\newblock Fast and accurate deep network learning by exponential linear units
  (elus).
\newblock In Yoshua Bengio and Yann LeCun, editors, {\em 4th International
  Conference on Learning Representations, {ICLR} 2016, San Juan, Puerto Rico,
  May 2-4, 2016, Conference Track Proceedings}, 2016.

\bibitem{9349197}
Matthias Delange, Rahaf Aljundi, Marc Masana, Sarah Parisot, Xu Jia, Ales
  Leonardis, Greg Slabaugh, and Tinne Tuytelaars.
\newblock A continual learning survey: Defying forgetting in classification
  tasks.
\newblock {\em IEEE Transactions on Pattern Analysis and Machine Intelligence},
  pages 1--1, 2021.

\bibitem{DBLP:conf/naacl/DevlinCLT19}
Jacob Devlin, Ming{-}Wei Chang, Kenton Lee, and Kristina Toutanova.
\newblock {BERT:} pre-training of deep bidirectional transformers for language
  understanding.
\newblock In Jill Burstein, Christy Doran, and Thamar Solorio, editors, {\em
  Proceedings of the 2019 Conference of the North American Chapter of the
  Association for Computational Linguistics: Human Language Technologies,
  {NAACL-HLT} 2019, Minneapolis, MN, USA, June 2-7, 2019, Volume 1 (Long and
  Short Papers)}, pages 4171--4186. Association for Computational Linguistics,
  2019.

\bibitem{DBLP:conf/iclr/DosovitskiyB0WZ21}
Alexey Dosovitskiy, Lucas Beyer, Alexander Kolesnikov, Dirk Weissenborn,
  Xiaohua Zhai, Thomas Unterthiner, Mostafa Dehghani, Matthias Minderer, Georg
  Heigold, Sylvain Gelly, Jakob Uszkoreit, and Neil Houlsby.
\newblock An image is worth 16x16 words: Transformers for image recognition at
  scale.
\newblock In {\em 9th International Conference on Learning Representations,
  {ICLR} 2021, Virtual Event, Austria, May 3-7, 2021}. OpenReview.net, 2021.

\bibitem{podnet}
Arthur Douillard, Matthieu Cord, Charles Ollion, Thomas Robert, and Eduardo
  Valle.
\newblock Podnet: Pooled outputs distillation for small-tasks incremental
  learning.
\newblock In {\em Proceedings of the IEEE European Conference on Computer
  Vision (ECCV)}, 2020.

\bibitem{DBLP:journals/corr/abs-2111-11326}
Arthur Douillard, Alexandre Ram{\'{e}}, Guillaume Couairon, and Matthieu Cord.
\newblock Dytox: Transformers for continual learning with dynamic token
  expansion.
\newblock {\em CoRR}, abs/2111.11326, 2021.

\bibitem{he2016deep}
Kaiming He, Xiangyu Zhang, Shaoqing Ren, and Jian Sun.
\newblock Deep residual learning for image recognition.
\newblock In {\em Proceedings of the IEEE conference on computer vision and
  pattern recognition}, 2016.

\bibitem{kd}
Geoffrey~E. Hinton, Oriol Vinyals, and Jeffrey Dean.
\newblock Distilling the knowledge in a neural network.
\newblock {\em CoRR}, abs/1503.02531, 2015.

\bibitem{Hou_2019_CVPR}
Saihui Hou, Xinyu Pan, Chen~Change Loy, Zilei Wang, and Dahua Lin.
\newblock Learning a unified classifier incrementally via rebalancing.
\newblock In {\em The IEEE Conference on Computer Vision and Pattern
  Recognition (CVPR)}, June 2019.

\bibitem{ewc}
James Kirkpatrick, Razvan Pascanu, Neil~C. Rabinowitz, Joel Veness, Guillaume
  Desjardins, Andrei~A. Rusu, Kieran Milan, John Quan, Tiago Ramalho, Agnieszka
  Grabska-Barwinska, Demis Hassabis, Claudia Clopath, Dharshan Kumaran, and
  Raia Hadsell.
\newblock Overcoming catastrophic forgetting in neural networks.
\newblock {\em Proceedings of the National Academy of Sciences}, 2017.

\bibitem{dirichlet}
Soochan Lee, Junsoo Ha, Dongsu Zhang, and Gunhee Kim.
\newblock A neural dirichlet process mixture model for task-free continual
  learning.
\newblock In {\em International Conference on Learning Representations}, 2020.

\bibitem{imm}
Sang-Woo Lee, Jin-Hwa Kim, Jaehyun Jun, Jung-Woo Ha, and Byoung-Tak Zhang.
\newblock Overcoming catastrophic forgetting by incremental moment matching.
\newblock In {\em Advances in neural information processing systems}, 2017.

\bibitem{li2022technical}
Duo Li, Guimei Cao, Yunlu Xu, Zhanzhan Cheng, and Yi Niu.
\newblock Technical report for iccv 2021 challenge sslad-track3b: Transformers
  are better continual learners.
\newblock {\em arXiv preprint arXiv:2201.04924}, 2022.

\bibitem{lwf}
Zhizhong Li and Derek Hoiem.
\newblock Learning without forgetting.
\newblock {\em IEEE Transactions on Pattern Analysis and Machine Intelligence},
  40(12):2935--2947, 2018.

\bibitem{liu2020generative}
Xialei Liu, Chenshen Wu, Mikel Menta, Luis Herranz, Bogdan Raducanu, Andrew~D
  Bagdanov, Shangling Jui, and Joost van~de Weijer.
\newblock Generative feature replay for class-incremental learning.
\newblock In {\em Conference on Computer Vision and Pattern Recognition
  Workshops}, 2020.

\bibitem{Liu2021SwinTH}
Ze Liu, Yutong Lin, Yue Cao, Han Hu, Yixuan Wei, Zheng Zhang, Stephen Lin, and
  Baining Guo.
\newblock Swin transformer: Hierarchical vision transformer using shifted
  windows.
\newblock {\em Proc. ICCV}, 2021.

\bibitem{liu2021swin}
Ze Liu, Yutong Lin, Yue Cao, Han Hu, Yixuan Wei, Zheng Zhang, Stephen Lin, and
  Baining Guo.
\newblock Swin transformer: Hierarchical vision transformer using shifted
  windows.
\newblock In {\em Proceedings of the IEEE/CVF International Conference on
  Computer Vision}, 2021.

\bibitem{gem}
David Lopez-Paz and Marc\textquotesingle~Aurelio Ranzato.
\newblock Gradient episodic memory for continual learning.
\newblock In {\em Advances in Neural Information Processing Systems 30}. 2017.

\bibitem{Maas13rectifiernonlinearities}
Andrew~L. Maas, Awni~Y. Hannun, and Andrew~Y. Ng.
\newblock Rectifier nonlinearities improve neural network acoustic models.
\newblock In {\em in ICML Workshop on Deep Learning for Audio, Speech and
  Language Processing}, 2013.

\bibitem{DBLP:journals/ijon/MaiLJQKS22}
Zheda Mai, Ruiwen Li, Jihwan Jeong, David Quispe, Hyunwoo Kim, and Scott
  Sanner.
\newblock Online continual learning in image classification: An empirical
  survey.
\newblock {\em Neurocomputing}, 469:28--51, 2022.

\bibitem{Mai2022OnlineCL}
Zheda Mai, Ruiwen Li, Jihwan Jeong, David Quispe, Hyunwoo~J. Kim, and Scott
  Sanner.
\newblock Online continual learning in image classification: An empirical
  survey.
\newblock {\em Neurocomputing}, 2022.

\bibitem{mallya2018packnet}
Arun Mallya and Svetlana Lazebnik.
\newblock Packnet: Adding multiple tasks to a single network by iterative
  pruning.
\newblock In {\em Proceedings of the IEEE Conference on Computer Vision and
  Pattern Recognition}, 2018.

\bibitem{masana2020class}
Marc Masana, Xialei Liu, Bartlomiej Twardowski, Mikel Menta, Andrew~D Bagdanov,
  and Joost van~de Weijer.
\newblock Class-incremental learning: survey and performance evaluation.
\newblock {\em arXiv preprint arXiv:2010.15277}, 2020.

\bibitem{mirzadeh2022architecture}
Seyed~Iman Mirzadeh, Arslan Chaudhry, Dong Yin, Timothy Nguyen, Razvan Pascanu,
  Dilan Gorur, and Mehrdad Farajtabar.
\newblock Architecture matters in continual learning.
\newblock {\em arXiv preprint arXiv:2202.00275}, 2022.

\bibitem{Nair2010RectifiedLU}
Vinod Nair and Geoffrey~E. Hinton.
\newblock Rectified linear units improve restricted boltzmann machines.
\newblock In {\em ICML}, 2010.

\bibitem{DBLP:journals/nn/ParisiKPKW19}
German~Ignacio Parisi, Ronald Kemker, Jose~L. Part, Christopher Kanan, and
  Stefan Wermter.
\newblock Continual lifelong learning with neural networks: {A} review.
\newblock {\em Neural Networks}, 2019.

\bibitem{pascanu2013revisiting}
Razvan Pascanu and Yoshua Bengio.
\newblock Revisiting natural gradient for deep networks.
\newblock {\em arXiv preprint arXiv:1301.3584}, 2013.

\bibitem{paul2021vision}
Sayak Paul and Pin-Yu Chen.
\newblock Vision transformers are robust learners.
\newblock {\em arXiv preprint arXiv:2105.07581}, 2021.

\bibitem{rebuffi2017icarl}
Sylvestre-Alvise Rebuffi, Alexander Kolesnikov, Georg Sperl, and Christoph~H
  Lampert.
\newblock icarl: Incremental classifier and representation learning.
\newblock In {\em Proceedings of the IEEE conference on Computer Vision and
  Pattern Recognition}, pages 2001--2010, 2017.

\bibitem{laplace}
Hippolyt Ritter, Aleksandar Botev, and David Barber.
\newblock Online structured laplace approximations for overcoming catastrophic
  forgetting.
\newblock In {\em Advances in Neural Information Processing Systems}, 2018.

\bibitem{DBLP:journals/tacl/RoySVG21}
Aurko Roy, Mohammad Saffar, Ashish Vaswani, and David Grangier.
\newblock Efficient content-based sparse attention with routing transformers.
\newblock {\em Trans. Assoc. Comput. Linguistics}, 9:53--68, 2021.

\bibitem{rusu2016progressive}
Andrei~A Rusu, Neil~C Rabinowitz, Guillaume Desjardins, Hubert Soyer, James
  Kirkpatrick, Koray Kavukcuoglu, Razvan Pascanu, and Raia Hadsell.
\newblock Progressive neural networks.
\newblock {\em arXiv preprint arXiv:1606.04671}, 2016.

\bibitem{DBLP:journals/corr/RusuRDSKKPH16}
Andrei~A. Rusu, Neil~C. Rabinowitz, Guillaume Desjardins, Hubert Soyer, James
  Kirkpatrick, Koray Kavukcuoglu, Razvan Pascanu, and Raia Hadsell.
\newblock Progressive neural networks.
\newblock {\em CoRR}, abs/1606.04671, 2016.

\bibitem{Sodhani2020TowardTR}
Shagun Sodhani, Sarath Chandar, and Yoshua Bengio.
\newblock Toward training recurrent neural networks for lifelong learning.
\newblock {\em Neural Computation}, 2020.

\bibitem{Touvron2021TrainingDI}
Hugo Touvron, Matthieu Cord, Matthijs Douze, Francisco Massa, Alexandre
  Sablayrolles, and Herv'e J'egou.
\newblock Training data-efficient image transformers \& distillation through
  attention.
\newblock In {\em ICML}, 2021.

\bibitem{DBLP:conf/nips/VaswaniSPUJGKP17}
Ashish Vaswani, Noam Shazeer, Niki Parmar, Jakob Uszkoreit, Llion Jones,
  Aidan~N. Gomez, Lukasz Kaiser, and Illia Polosukhin.
\newblock Attention is all you need.
\newblock In Isabelle Guyon, Ulrike von Luxburg, Samy Bengio, Hanna~M. Wallach,
  Rob Fergus, S.~V.~N. Vishwanathan, and Roman Garnett, editors, {\em Advances
  in Neural Information Processing Systems 30: Annual Conference on Neural
  Information Processing Systems 2017, December 4-9, 2017, Long Beach, CA,
  {USA}}, 2017.

\bibitem{Wang2021EndtoEndVI}
Yuqing Wang, Zhaoliang Xu, Xinlong Wang, Chunhua Shen, Baoshan Cheng, Hao Shen,
  and Huaxia Xia.
\newblock End-to-end video instance segmentation with transformers.
\newblock {\em 2021 IEEE/CVF Conference on Computer Vision and Pattern
  Recognition (CVPR)}, 2021.

\bibitem{early_conv}
Tete Xiao, Mannat Singh, Eric Mintun, Trevor Darrell, Piotr Doll{\'{a}}r, and
  Ross~B. Girshick.
\newblock Early convolutions help transformers see better.
\newblock {\em CoRR}, abs/2106.14881, 2021.

\bibitem{DBLP:conf/nips/XuZ18}
Ju Xu and Zhanxing Zhu.
\newblock Reinforced continual learning.
\newblock In Samy Bengio, Hanna~M. Wallach, Hugo Larochelle, Kristen Grauman,
  Nicol{\`{o}} Cesa{-}Bianchi, and Roman Garnett, editors, {\em Advances in
  Neural Information Processing Systems 31: Annual Conference on Neural
  Information Processing Systems 2018, NeurIPS 2018, December 3-8, 2018,
  Montr{\'{e}}al, Canada}, 2018.

\bibitem{Yang2021FocalSF}
Jianwei Yang, Chunyuan Li, Pengchuan Zhang, Xiyang Dai, Bin Xiao, Lu Yuan, and
  Jianfeng Gao.
\newblock Focal self-attention for local-global interactions in vision
  transformers.
\newblock {\em ArXiv}, abs/2107.00641, 2021.

\bibitem{DBLP:conf/iclr/YoonYLH18}
Jaehong Yoon, Eunho Yang, Jeongtae Lee, and Sung~Ju Hwang.
\newblock Lifelong learning with dynamically expandable networks.
\newblock In {\em 6th International Conference on Learning Representations,
  {ICLR} 2018, Vancouver, BC, Canada, April 30 - May 3, 2018, Conference Track
  Proceedings}. OpenReview.net, 2018.

\bibitem{yu2020semantic}
Lu Yu, Bartlomiej Twardowski, Xialei Liu, Luis Herranz, Kai Wang, Yongmei
  Cheng, Shangling Jui, and Joost van~de Weijer.
\newblock Semantic drift compensation for class-incremental learning.
\newblock In {\em Proceedings of the IEEE/CVF Conference on Computer Vision and
  Pattern Recognition}, pages 6982--6991, 2020.

\bibitem{DBLP:journals/corr/abs-2112-06103}
Pei Yu, Yinpeng Chen, Ying Jin, and Zicheng Liu.
\newblock Improving vision transformers for incremental learning.
\newblock {\em CoRR}, abs/2112.06103, 2021.

\bibitem{si}
Friedemann Zenke, Ben Poole, and Surya Ganguli.
\newblock Continual learning through synaptic intelligence.
\newblock In Doina Precup and Yee~Whye Teh, editors, {\em Proceedings of the
  34th International Conference on Machine Learning}, Proceedings of Machine
  Learning Research, 2017.

\bibitem{Zenke2017ContinualLT}
Friedemann Zenke, Ben Poole, and Surya Ganguli.
\newblock Continual learning through synaptic intelligence.
\newblock {\em Proceedings of machine learning research}, 2017.

\end{thebibliography}
}

\clearpage


\appendix

\section{Additional Settings}
\label{tag_appendix}

We experiment on two further CIFAR100 settings with distinct cardinality of base task classes:
\begin{itemize}
    \item CIFAR100/20 Base, with 20 base task classes followed by 8 incremental tasks with 10 classes each,
    \item CIFAR100/50 Base, with 50 base task classes followed by 5 incremental tasks with 10 classes each.
\end{itemize}

\begin{figure}[h!]
\centering
\centerline{\includegraphics[width=\columnwidth]{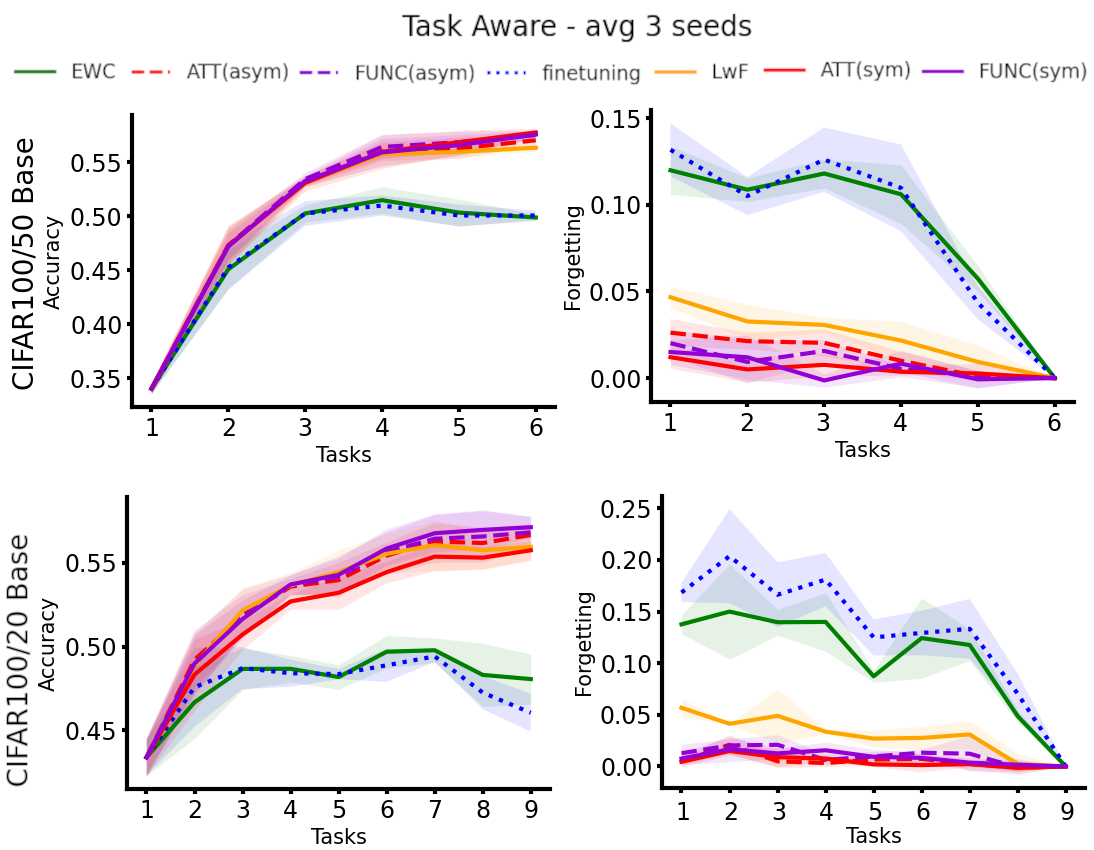}}
\caption{Mean and standard deviation of task-aware accuracy and forgetting scores for the additional CIFAR100/20 and CIFAR100/50 settings (over 3 random runs).}
\label{fig:cifar100_ttaw_50_20}
\end{figure}

\begin{figure}[h!]
\begin{center}
\centerline{\includegraphics[width=\columnwidth]{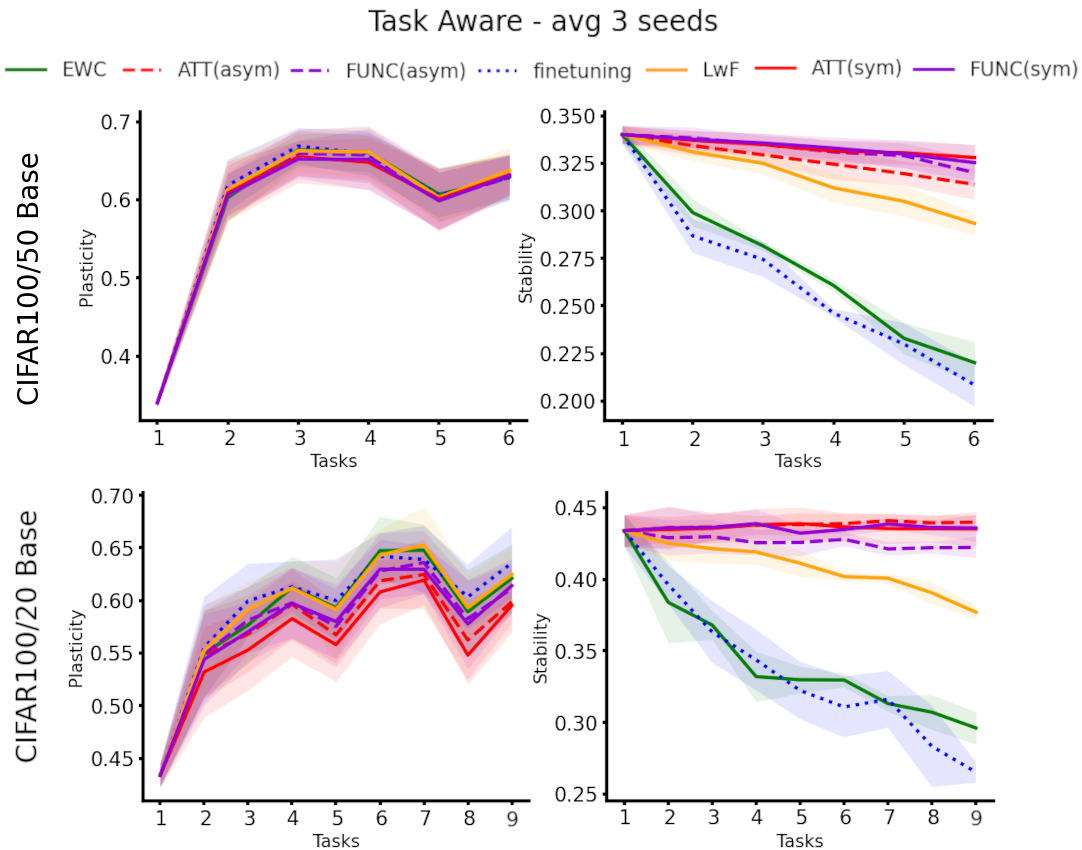}}
\caption{Mean and standard deviation of task-aware plasticity-stability scores for the additional CIFAR100/20 and CIFAR100/50 settings (over 3 random runs).}
\label{fig:cifar100_other_taw}
\end{center}
\vskip -0.4in
\end{figure}

The task aware accuracy and forgetting scores on these are shown in Figure \ref{fig:cifar100_ttaw_50_20}. We find the PAD-based losses to consistently outperform other regularization approaches with LwF being the closest tie. Along the direction of plasticity-stability trade off (see Figure \ref{fig:cifar100_other_taw}), we observe that: (a) the attention-based $\mathcal{L}_\text{(a)sym}$ losses retain better rigidity than their functional counterparts, and (b) the asymmetric variants are more plastic than their symmetric counterparts across these settings. These trends further validate our hypotheses in sections \ref{sec:att_reg} and \ref{sec:asym_reg}, respectively.

\begin{figure*}[h!]
\centering
\centerline{\includegraphics[width=1\textwidth]{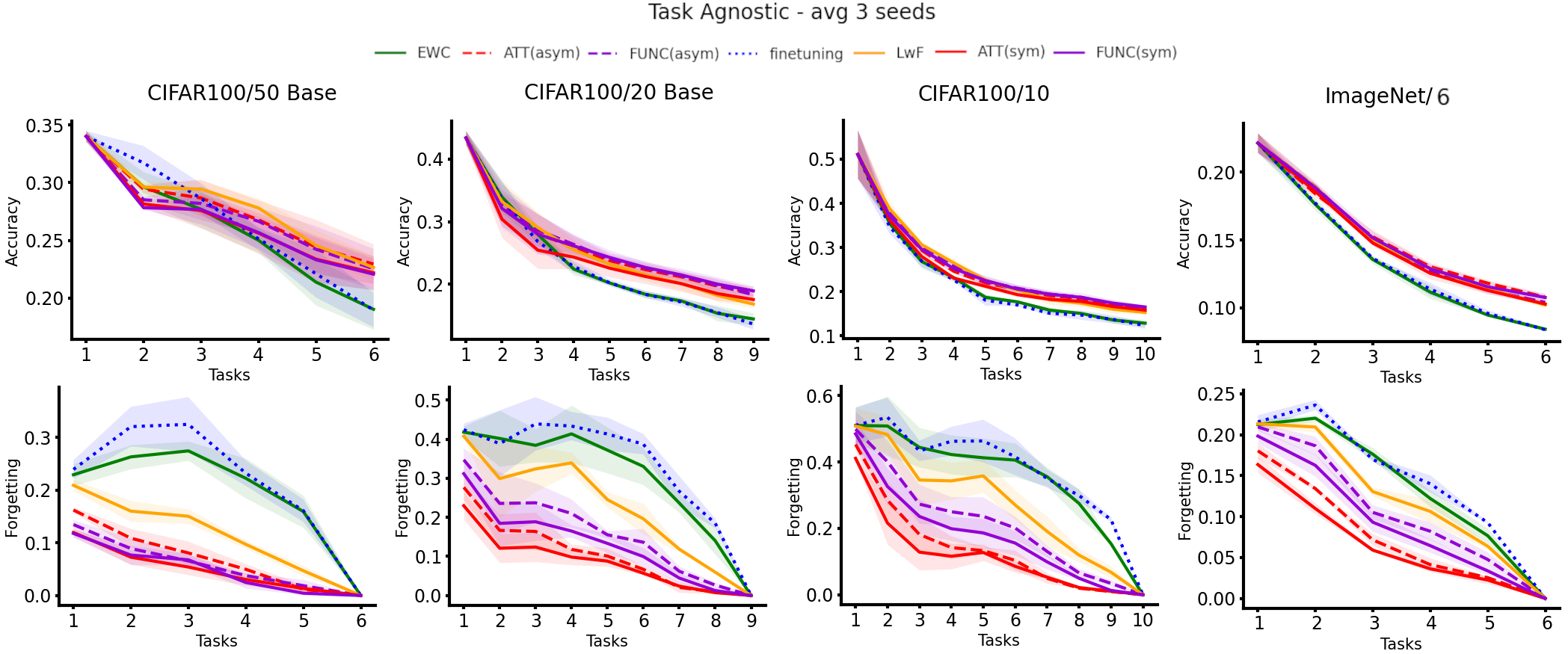}}
\caption{Mean and standard deviation of task-agnostic accuracy and forgetting scores for CIFAR100/10, CIFAR100/20, CIFAR100/50, and ImageNet/6 settings (over 3 random runs). The larger proportion of base task classes (for example, CIFAR100/50) gives rise to higher variations of accuracies and lower variation of forgetting scores across methods -- with the latter indicating the inclination of ViTs towards better generalization and preservation of knowledge.}
\label{fig:cifar100_tag_settings}
\end{figure*}

\section{Task Agnostic Results}

Figure \ref{fig:cifar100_tag_settings} depicts the task-agnostic accuracy and forgetting scores for the settings mentioned in the main paper as well as in Appendix \ref{tag_appendix}. Given the contradictory terms of resource-scarce \textit{exemplar-free} CL and data-hungry ViTs, task-agnostic evaluations can be seen to be particularly challenging. The further avoidance of heavier data augmentations in our training settings give rise to two major repercussions across the task-agnostic accuracies: (a) the scores remain consistently low, and (b) the models show smaller yet consistent variations in performances across all settings. 

That said, we find functional $\mathcal{L}_\text{(a)sym}$ losses to be performing the best on all but CIFAR100/50 setting. The larger proportion of base task classes in the latter setting can be seen to be greatly benefiting the learning of LwF (the least parameterized loss term). Further on the class proportions, we observe that an equal spread of classes across the tasks can be seen to have a smoothing effect on the variations of scores across different methods.

 On the contrary, the CIFAR100/50 setting leads to low variability of task-agnostic forgetting scores across the methods. This can again be attributed to the fact that a larger first task better leverages the generalization capabilities of ViTs thus advancing them at avoiding forgetting  over the subsequent incremental steps. This further adds to our reasoning regarding the natural resilience of ViTs to CL settings. When compared across methods, the attentional variants of $\mathcal{L}_\text{(a)sym}$ can be seen to display the least amount of forgetting followed by their functional counterparts.




\end{document}